\documentclass{article}
\usepackage{amsmath,amssymb,amsthm}
    \usepackage[preprint, nonatbib]{neurips_2020}

\usepackage[utf8]{inputenc} % allow utf-8 input
\usepackage[T1]{fontenc}    % use 8-bit T1 fonts
\usepackage{hyperref}       % hyperlinks
\usepackage{url}            % simple URL typesetting
\usepackage{booktabs}       % professional-quality tables
\usepackage{amsfonts}       % blackboard math symbols
\usepackage{nicefrac}       % compact symbols for 1/2, etc.
\usepackage{microtype}      % microtypography
\usepackage{graphicx} 
\usepackage{multirow}
\usepackage{subcaption}
\usepackage{caption}

\usepackage[style=numeric,sorting=none]{biblatex}

\usepackage{mathrsfs}

\title{Predicting Generalization in Deep Learning via \\ Local Measures of Distortion}
\author{
    Abhejit~Rajagopal \\
    University of California, San Francisco\\
    \texttt{abhejit.rajagopal@ucsf.edu} \\
    \And
    Vamshi C. Madala \\
    Universtiy of California, Santa Barbara \\
    \texttt{vamshichowdary@ucsb.edu} \\
    \And
    Shivkumar~Chandrasekaran \\
    University of California, Santa Barbara\\
    \texttt{shiv@ucsb.edu} \\
    \And
    Peder~E.~Z.~Larson \\
    Universtiy of California, San Francisco \\
    \texttt{peder.larson@ucsf.edu} \\
}
\begin{document}
\maketitle
\vspace{-4mm}
\begin{abstract}
\vspace{-2mm}
We study generalization in deep learning by appealing to complexity measures originally developed in approximation and information theory. While these concepts are challenged by the high-dimensional and data-defined nature of deep learning, we show that simple vector quantization approaches such as PCA, GMMs, and SVMs capture their spirit when applied layer-wise to deep extracted features giving rise to relatively inexpensive complexity measures that correlate well with generalization performance. We discuss our results in 2020 NeurIPS PGDL challenge.
\end{abstract}

% \vfill
\vspace{-5mm}
\section{Background}
\vspace{-2mm}
In approximation theory, generalization error is measured by the maximum error between an approximation $f^*(x)$ and the true target function $f(x)$ in a domain $\mathcal{D}$, as $\min_{x \in \mathcal{D}} ||f(x)-f^*(x)||_\infty$. For classification problems, this definition is often relaxed to estimate how often the target function is matched over the whole domain, e.g.~using a normalized 0-norm to measure train/test classification ``accuracy'', or a more-continuous $p$-norm to gauge average train/test error. In any case, measuring the true generalization performance requires knowledge of the target function over the whole domain, which presents an issue for deep learning problems with finite high-dimensional data.

Based on recent works~\cite{jiang2018predicting,jiang2019fantastic}, however, the \textit{2020 NeurIPS PGDL Challenge} seeks predictors of deep neural network~(DNN) generalization performance that are based solely on training data, DNN parameters, and training hyperparameters~(e.g.~learning rate, optimizer, training loss value, etc). Here, the true generalization is measured as the accuracy of the model on a large withheld set of test data.
For each architecture (e.g.~fully-convolutional or CNN + fully-connected) and task (e.g.~image classification on CIFAR10 or SVNH), the challenge provides several trained networks with comparable training performance ($\geq 90\%$) but significantly different testing performance.
The task is then to develop a method for predicting the generalization error for a withheld set of networks trained for the same or new tasks with the same or different hyperparameters.

Our approach to this challenge is based on recent work in approximation theory~\cite{mhaskar2016deep,mhaskar2020analysis,rajagopal2019high}, which suggests that deep networks constructed using the true compositional structure of the target function can not only escape the curse of dimensionality but also converge to the target with provably good generalization error by penalizing the roughness of the interpolant, even when only given point-data. For approximation from point data, there is an inherent trade-off between (a)~fitting all the training points, (b)~the order or complexity of the approximation, and (c)~the rate of convergence.
Thus, for a given parameterization (DNN with weights), we seek to predict its generalization performance using measures of the model's \textit{regularity or smoothness}. While a comprehensive approach would count oscillations at every node~(e.g.~neuron) in the feedforward directed acyclic graph $\mathcal{G}$ shared by the target function and its neural network approximation, due to the large nodal degree of typical DNNs and for the purpose of this challenge, we seek a less expensive measure of the network's smoothness that can operate within the challenge's computational parameters ($\leq 5$min/model, $\leq 20$GB RAM).

\newpage
\section{Measures of Distortion}
% \vspace{-2mm}
Vector quantization (VQ) is intimately connected to both statistical learning theory and pattern recognition. In communications, VQ is used at the receiver to determine which (possibly high-dimensional) codewords were transmitted. If we view image classification through the same lens, we see that deep networks are simply an elaborate decoder for the classification labels (e.g.~one-hot encoded or multi-level). For deterministic feedforward networks that are trained well, features extracted in the intermediate layers are drop-in replacements for the inputs. That is, the role of the deep network is to degeneratively transform seemingly disparate, high dimensional image vectors to \textit{clustered} representations that are ultimately quantized to the classification labels.

Although the final quantization rule can be quite complicated due to the high nodal degree
% $d_v$
and nonlinear activation of many DNNs,
% the Theorem~\ref{theorem:deepMSN}
\cite{rajagopal2019high}
provides the insight that for valid compositional architectures (decoders), minimizing the Sobolev norm of the nodal functions yields provably good performance. This implies that for two networks with the same training performance, the network with smoother nodes/layers will tend to have better generalization performance as the number of samples $N \rightarrow \infty$. 
% One way to capture the smoothness of a given layer is by computing the complexity of the required VQ rule.
We postulate that a computationally efficient alternative to computing the Sobolev norm at each node is to instead estimate the complexity of the required VQ rule at that layer.
For example, if we view the feature distribution of training data at an intermediate layer $\ell$ as a symbol constellation, a complicated VQ rule would correspond to a highly oscillatory (rough) function at the next layer $\ell+1$.

\subsection{Symbol Distortion and Euclidean Metrics}
\begin{figure}[hbt!]%
    \vspace{-2mm}
    %\centering
    \subfloat{{\includegraphics[width=14cm]{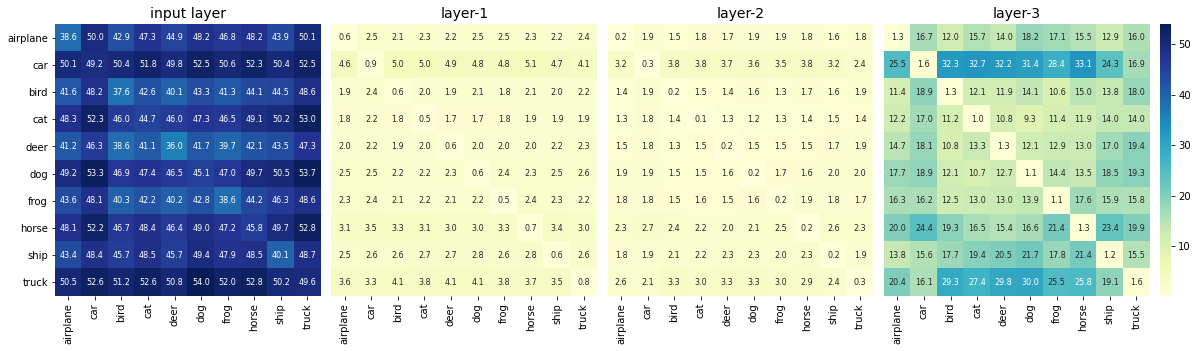} }}%
    \vspace{-2mm}
    \caption{$L_2$ distortion matrices as a function of layer (PGDL \texttt{Model 668}). 
    The diagonal and off-diagonal elements represent the mean minimum intra- and inter-class $2$-norm distance, respectively.
    }%
    \label{fig:mesh}%
    \vspace{-6mm}
\end{figure}

\vspace{-2mm}
We can use $p$-norms to measure the expected complexity or distortion of VQ at an intermediate layer, starting with the input layer (prior to any network operation). This is commonly used in VQ to measure the mismatch between the source symbols and reconstructed symbols~\cite{seo2003soft}, and also has connections to the mesh norm $\min_{x,y \in T, x \neq y} \|x - y \|$ that is central to approximation theory. As seen in Figure~\ref{fig:mesh}, the mean minimum intra-class distance of training data reduces deeper in the network.

\subsection{Label Distortion}
\vspace{-2mm}
Due to the high nodal degree of current DNNs, $p$-norms may systematically over estimate the complexity of the required VQ. A simpler metric closer to the classification task is the accuracy of a nearest-neighbor-type classifier. Good networks will have diagonally dominant confusion matrices, corresponding to simple classification boundaries even if the distance between samples is large. However, such classifiers also depend on $p$ norms. To better capture distances between the high dimensional feature vectors, we can use dimensionality reduction techniques, e.g.~kernel PCA.

\subsubsection{Kernel Principal Component Analysis (kPCA) and Gaussian Mixture Models (GMMs)}
\vspace{-2mm}
Unfortunately, computing the accuracy of a nearest-neighbor classifier using transformed coordinates of the training data's features does not correlate well with generalization (test) performance of models with good training accuracy, although it can help identify networks with unwanted degeneracy. Instead, we compute the principal components and evaluate the ``validation'' accuracy of the resulting nearest neighbor classifier using different mutually-exclusive subsets of the training data. 
% That is, we first split the training data into several mutually-exclusive subsets, and perform kernel PCA (kPCA) on each one independently. For each kPCA model, we evaluate the performance of 
Since each kernel PCA model is trained on only a subset of the training data's features (e.g.~at layer $\ell$), the resulting confusion matrix measures how disparate the features of training data are, or how generalizable the features are across different subsets.

To enable faster computation during the competition, we additionally employ Gaussian mixture models (GMMs) that are initialized and trained on each class independently with 3-5 components per class. In addition to enabling a fast classification computation (e.g.~closest mixture centroid), GMMs can be used to measure the confidence in each prediction with respect to the distribution of the chosen training subset. That is, for each class $i$, we compute the distortion $d_i = \mathbb{E}_{x_j} \{\max_{k \in GMM_j} p(\Theta_k | x_i)\} \; \forall \; j$ where $j$ represents each class in the dataset; $k$ corresponds to each of the GMM components of class $j$ represented by $GMM_j$; and the expectation is approximated by calculating the mean of inner maximum values for all data points in class $j$.

\vspace{-2mm}
\subsection{Support Vector Machines (SVMs)}
\vspace{-2mm}
To provide additional supervision into the classifier, we can employ support vector machines (SVMs). Unlike a nearest-neighbor or GMM-based classifiers, kernel SVMs can easily achieve arbitrary training accuracy, approaching the performance of neural network layers at the expense of more parameters. That is, in addition to computing confusion matrices and confidence measures based on the distance to the boundary (e.g.~margin error), the complexity of the SVM can be directly measured using the number of support vectors required for $\epsilon$ training error. The more complicated the decision function, the more support vectors that are required, and higher the expected Sobolev norm.

\vspace{-2mm}
\section{Results}
\begin{figure}[hbt!]
\begin{subfigure}{\linewidth}
\centering
\includegraphics[width=13.5cm]{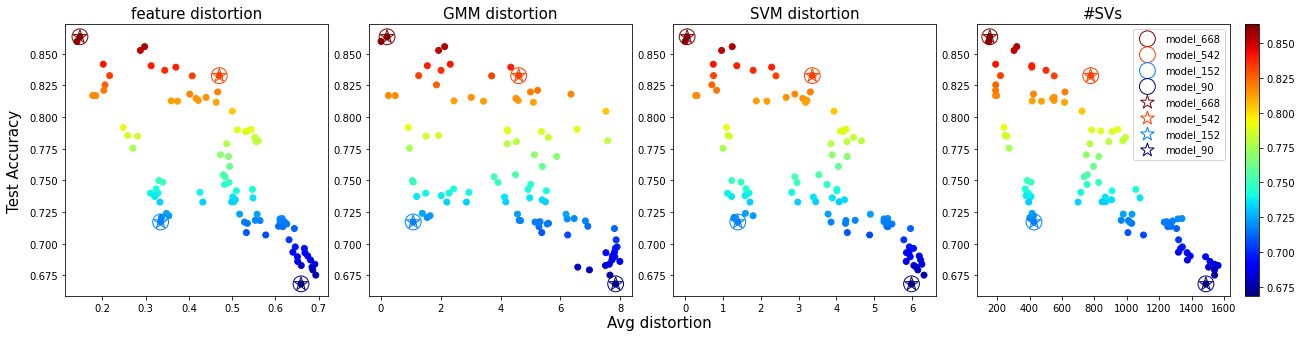}
\vspace{-2mm}
\caption{Task-1 models.}
\label{fig:corr_task1}
\end{subfigure}\\%[1ex]
\begin{subfigure}{\linewidth}
\centering
\includegraphics[width=13.5cm]{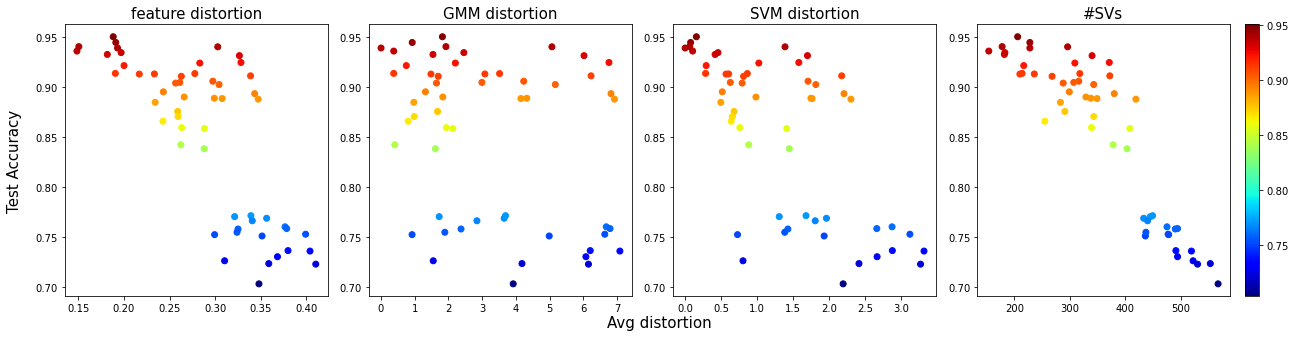}
\vspace{-2mm}
\caption{Task-2 models.}
\label{fig:corr_task2}
\end{subfigure}
\vspace{-2mm}
\caption{Distortion-test accuracy correlations for different distortion measures.}
\label{fig:corr_task_comp}
\end{figure}

\vspace{-2mm}
For the PGDL competition, we evaluated how well the aforementioned measures of distortion correlate with test performance. With the exception of the direct SVM complexity metric (number of support vectors), for each model we computed the layer-wise symbol and label distortion using intermediate features ($L_2$ distance), their projected (kPCA) coordinates relative to GMM centroids, and kernel (radial basis function) SVMs. The normalized trace (mean of the diagonal) of these matrices was used as the final complexity measure for these cases. The correlation with the test accuracy for each Task-1 and Task-2 DNN model in the competition's public dataset is shown in Figures~\ref{fig:corr_task1}-\ref{fig:corr_task2}, while a comparison of correlation values is displayed in Table~1. As can be seen, the number of support vectors (\#SVs) provides the best correlation for both Task-1 and Task-2 models.
\vspace{-2mm}

\vspace{-7mm}
\begin{table}[hbt!]
\centering
\small
\caption{$R^2$ values of distortion-test accuracy correlations}
\vspace{1mm}
%\begin{tabular}{|p{2cm}|p{2cm}|p{2cm}|p{2cm}|p{2cm}|}
\begin{tabular}{c|c|c|c|c}
%\hline
\multirow{2}{*}{\textbf{Distortion measure}} & \multicolumn{2}{c}{\textbf{Task-1}}
&
\multicolumn{2}{c}{\textbf{Task-2}} \\
\cline{2-3}\cline{4-5}
&\textbf{Trainset} & \textbf{Testset} & \textbf{Trainset} &\textbf{Testset} \\
\hline\hline
    Intermediate features & 0.53 & 0.61 & 0.56 & 0.9\\
\hline
    GMM & 0.31 & 0.39 & 0.10 & 0.43\\
\hline
    SVM & 0.50 & 0.71 & 0.44 & 0.93\\
\hline
    \#SVs & \textbf{0.59} & \textbf{0.93} & \textbf{0.81} & \textbf{0.98}\\
\hline
\end{tabular}
\label{table:r2_values}
\end{table}

\begin{figure}[hbt!]
\begin{subfigure}{\linewidth}
\centering
\includegraphics[width=13.5cm]{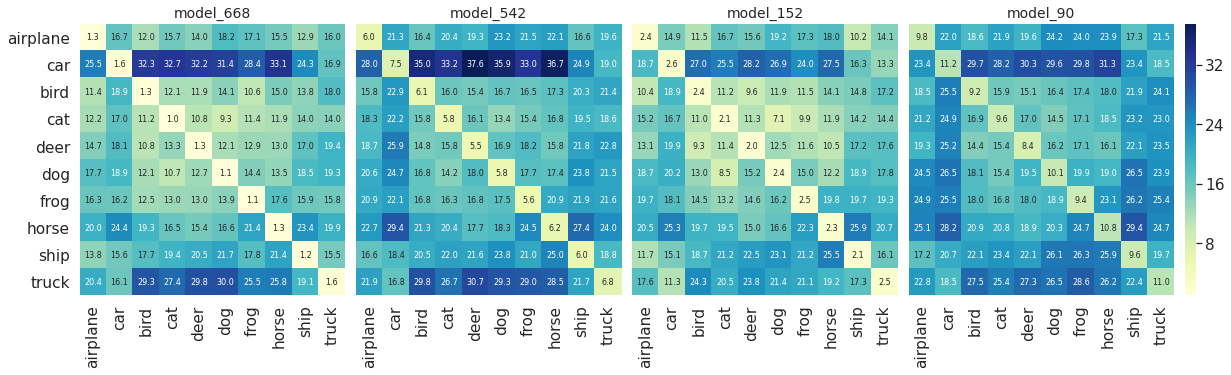}
\vspace{-2mm}
\caption{$L_2$ Distortion.}
\label{fig:dist_feat}
\end{subfigure}\\%[1ex]
\begin{subfigure}{\linewidth}
\centering
\includegraphics[width=13.5cm]{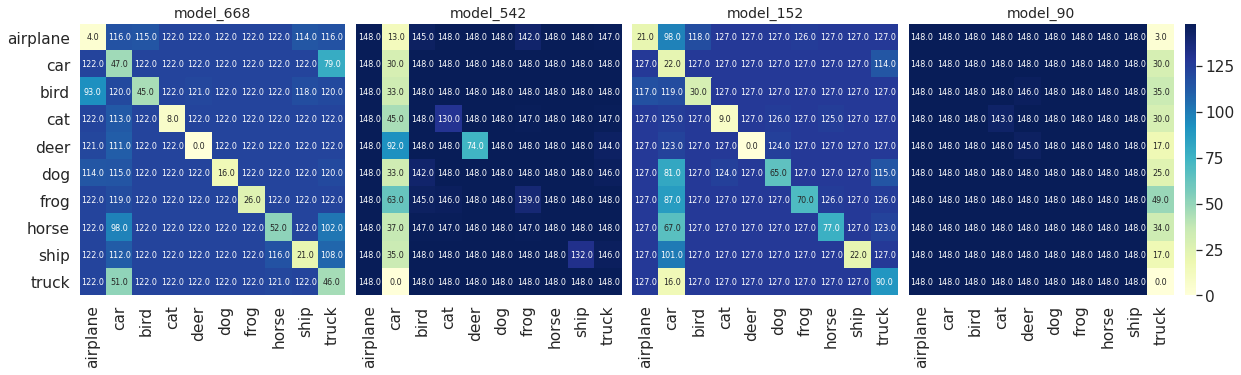}
\vspace{-2mm}
\caption{GMM label distortion.}
\label{fig:dist_gmm}
\end{subfigure}\\%[1ex]
\begin{subfigure}{\linewidth}
\centering
\includegraphics[width=13.5cm]{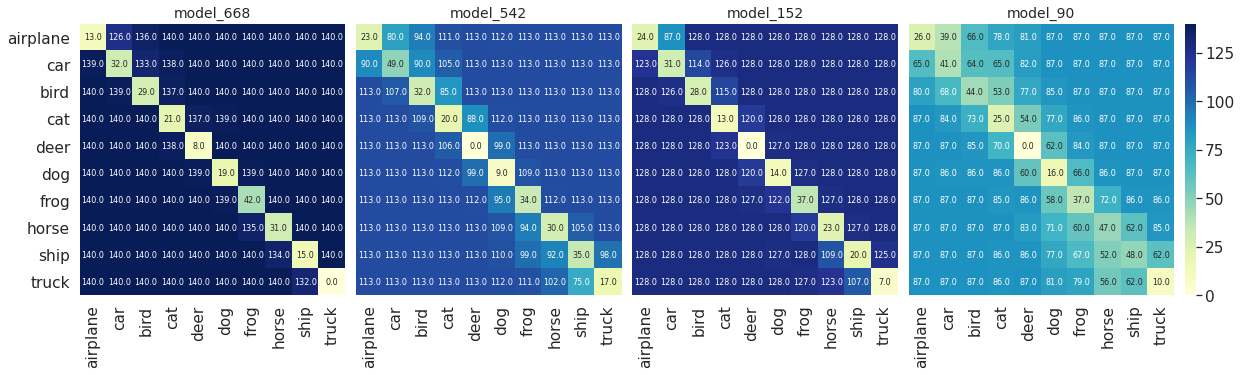}
\vspace{-2mm}
\caption{SVM label distortion.}
\label{fig:dist_svm}
\end{subfigure}
\vspace{-2mm}
\caption{Comparison of distortion measures.}
\label{fig:dist_comp}
\end{figure}

\newpage
To further demonstrate the properties of our complexity measures, we analyse the distortion matrices for four representative models from the Task~1 public dataset, indicated by stars in the correlation plots of Figure~\ref{fig:corr_task1}. Despite having similar training performance, these models differ in their generalization accuracy on the test set, as: model 668: 86.3\%; model 542: 83.3\%; model 152: 71.7\% and model 90: 66.8\%. The corresponding distortion matrices for these models are shown in Figure~\ref{fig:dist_comp}. While the $L_2$ distortion measure (Fig.~\ref{fig:dist_feat}) is able to distinguish a well generalized model from a poorly generalized one (by comparing the relative normalized trace), it is harder to draw a conclusion by just looking at a single model's distortion matrix. This may be attributed to the limitation of the $2$-norms in computing representative distances between high dimensional features, as previously mentioned. The SVM distortion measure overcomes this problem, as can be seen from Figure~\ref{fig:dist_svm}. SVM distortions are able to quantify the distances between the points within a class and also between the point clusters of different classes while maintaining a better correlation with model's generalization. 
\vspace{-4mm}

\begin{figure}[h!]%
    %\vspace{-10pt}
    %\centering
    \subfloat{{\includegraphics[width=13.5cm]{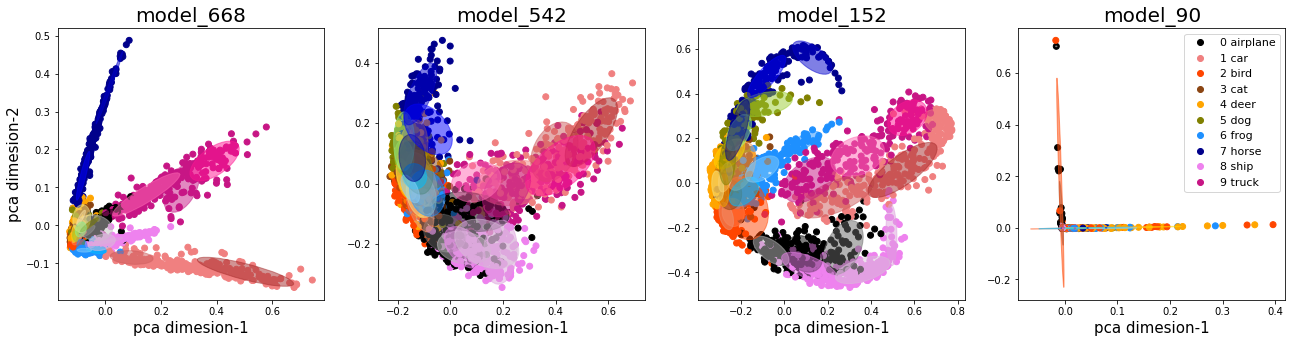} }}%
    \caption{Label wise GMM clusters for different models.}%
    \label{fig:gmm_clusters}%
    %\vspace{-10pt}
\end{figure}

\vspace{-2mm}
The kPCA + GMM distortion measure (Figure~\ref{fig:dist_gmm}) tends to approach this but fails to quantify the distortion for certain models. This might be due to the constraint imposed by the radial basis function (RBF) kernel used in kPCA, and a forced reduction in the dimensionality (from 100s to just 3). This can be seen from Figure~\ref{fig:gmm_clusters}, which depicts the feature vectors for the above models after applying kPCA and fitting GMMs. The ellipses of same color but with different gradients represent the clusters within a class. These plots provide a visual validation to our claim that better generalized models have smoother intermediate feature representations and thus better separated class clusters.

\newpage

\section{Conclusion}
\vspace{-2mm}
We showed that simple measures of regularity can be used to predict generalization, laying the foundation for exploring generalization in deep learning from the point of view of approximation theory and computational harmonic analysis. We developed simple metrics to measure the smoothness of DNN layers by exploiting the similarity between intermediate feature representations extracted by the models. We evaluated these metrics against VGG-like CNN models and Network-in-network models and demonstrated that these metrics correlate well with the models' generalization, \textit{independent of the training strategy}. Currently our proposed metrics do not take into account the notion of adversarial performance or architectural hyperparameters, and so in our future work we aim to incorporate these notions to develop more accurate complexity measures that better match theoretical results.

\begin{ack}
This research was supported by AFRL grant \#FA8650-18-C-1137, NIH/NCI grant \#R01CA229354, and NIH/NIBIB grant \#1F32EB030411-01.
\end{ack}

\printbibliography[title={References}]

\end{document}